\documentclass{article}
\usepackage{spconf,amsmath,graphicx}
\usepackage{colortbl}
\usepackage{enumitem}
\setlist{nosep, leftmargin=14pt}
\usepackage{overpic}
\usepackage{mwe} 
\usepackage{subcaption}
\usepackage{contour}
\contourlength{1pt}

\title{Fully Automatic Data Labeling for Ultrasound Screen Detection}
\name{%
  Alberto Gomez$^{\star}$ \qquad Jorge Oliveira \qquad Ramon Casero \qquad Agis Chartsias
}

\address{Ultromics Ltd, Oxford, UK}
\begin{document}

\maketitle

\begin{abstract}
Ultrasound (US) machines display images on a built-in monitor, but routine transfer to hospital systems relies on DICOM. We propose a fully automatic method to generate labeled data that can be used to train a screen detector model, and a pipeline to use that model to extract and rectify the US image from a photograph of the monitor, without any need for human annotation. This removes the DICOM bottleneck and enables rapid testing and prototyping of new algorithms. In a proof-of-concept study, the rectified images retained enough visual fidelity to classify cardiac views with a balanced accuracy of 0.79 with respect to the native DICOMs.
\end{abstract}

\section{Introduction}

Echocardiographic (echo) acquisition systems typically store echo images in DICOM format, and these images are made available to users on other devices via the internal hospital network. 
Differently to most other medical imaging modalities, echo acquisition systems are built with a screen, to allow for real-time guidance of the probe. 
Many applications would benefit from rapid, often real-time access to the data for further processing. To this end, some manufacturers provide dedicated communication protocols typically via a cable connecting the acquisition system and the analysis machine (e.g. HDMI), however a cable connection requires purposeful set-up and may be cumbersome and requires access to manufacturer protocol. We argue that capturing the content of the screen with something as simple as a video camera (handheld device) may allow seamless forwarding of the data for further processing, e.g. in mobile or augmented reality applications. 

Here we focus on the challenge of detecting the screen content from a picture of the screen (and the associated data labeling burden), correcting for perspective distortion to reshape the images to their original shape. 

\subsection{Related work}
Treivase et al \cite{treivase2020screen} investigated ultrasound (US) screen tracking with the aim of capturing and analyzing screen content, using patterned stickers placed on the corners of the screen and detecting those trackers. Aside from this work (which requires modifying the US system and collecting training data), most relevant literature is related to object detection and classification. Notably, a multi-task Unet was proposed in \cite{zhu2022multi} that performs both localized heatmaps and image-based classification, therefore suited to our problem of localizing the screen corners and detecting the presence of a screen in the image. 

\subsection{Contributions}

Building upon the concept in \cite{treivase2020screen} and using the architecture in \cite{zhu2022multi}, we present a novel strategy for screen detection and localization with the following contributions: (i) a method to generate self-annotated synthetic data, (ii) training a CNN with a multi-task loss to localize and detect the screen, (iii) preliminary evaluation in both synthetic and real data, and (iv) exemplifying downstream use by using the reconstructed images with a view classifier trained on standard echo images.

\begin{figure}[htb!]
\centering
\offinterlineskip
\begin{overpic}[height=0.41\linewidth]{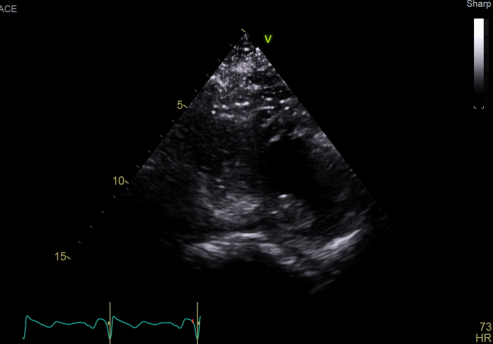}
\put(5,60){\color{white}\textbf{\large a}}
\end{overpic}%
\begin{overpic}[height=0.41\linewidth]{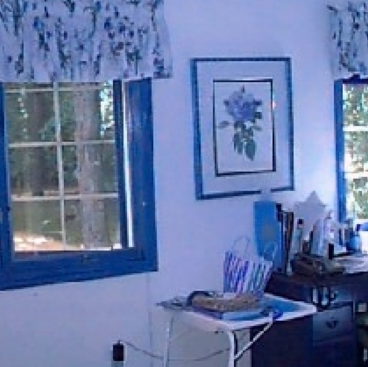}
\put(5,90){\color{white}\textbf{\large b}}
\end{overpic}\par
\begin{overpic}[height=0.41\linewidth]{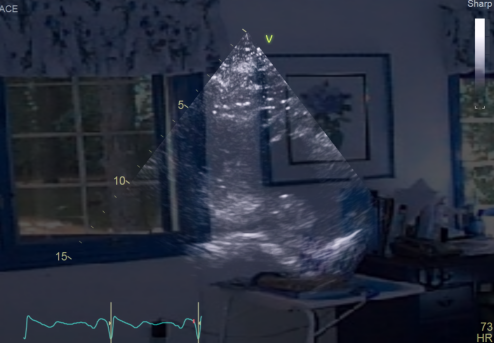}
\put(5,60){\color{white}\textbf{\large c}}
\end{overpic}%
\begin{overpic}[height=0.41\linewidth]{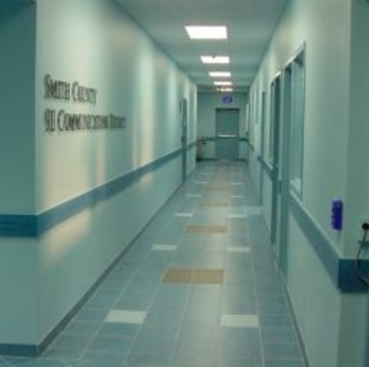}
\put(5,90){\color{white}\textbf{\large d}}
\end{overpic}\par
\begin{overpic}[width=0.5\linewidth]{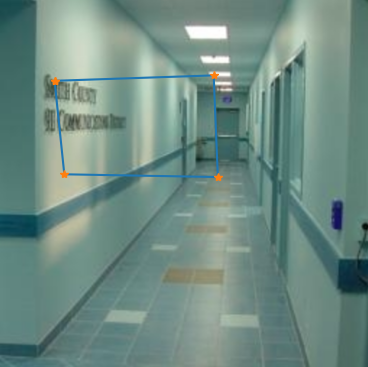}
\put(5,90){\color{white}\textbf{\large e}}
\end{overpic}%
\begin{overpic}[width=0.5\linewidth]{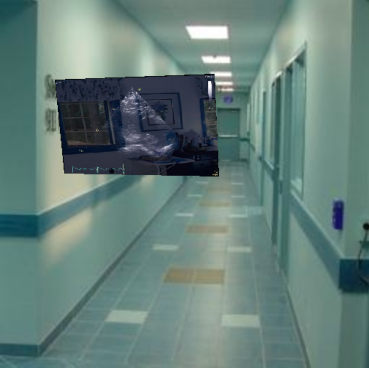}
\put(5,90){\color{white}\textbf{\large f}}
\end{overpic}
\caption{Steps in the creation of a synthetic image showing an echo screen with realistic reflection artifacts.}
\label{fig:synthetic-image-generation}
\end{figure}

\section{Materials and methods}

We propose a data preparation strategy to synthesize fully annotated data for screen detection without the need for human annotations,  a multi-task model to detect and localize the US screen, and a pipeline to extract image data from the detected screen. Each step is described in turn.

\subsection{Datasets}
We created a synthetic dataset of images that have an ultrasound screen on an indoor background, using two datasets: a background data set (MIT Indoors data set from CVPR2009 \cite{quattoni2009recognizing}, which contains images from 67 indoor categories), and a private anonymized US imaging dataset (with studies from 1000 adult patients with a variety of cardiac diseases throughout the USA, averaging 52 echo clips/study). We randomly split the 67 background categories into 50 for training, 12 for validation, and 5 for test. The US data was split by patient into training (75\%), validation (18\%) and test (7\%).

We compiled a ``real'' dataset by taking 100 pictures of ultrasound images displayed on a screen (a 12 inch tablet), against a variety of indoor backgrounds, from a diversity of view points and different amounts of reflections on the screen (manually labeled by picking the four corners), plus 100 pictures of indoor scenes. The real dataset was used for test only.

\subsection{Synthetic data generation}

Using the above datasets, we synthesize realistic natural images with a diversity of backgrounds on which a rectangular screen showing ultrasound images is visible, in a  random orientation, alongside the coordinates of the four corners of the screen. From the empirical observation that a major challenge in screen detection is reflection artifacts, we propose adding synthetic reflections to make the model more robust. 

To create synthetic reflections, we use screen blending \cite{Harrelson2024compositing}, where a reflection is blended into a picture as follows. Let $S$ be an echo image (Fig. \ref{fig:synthetic-image-generation}.a), and $R$ be the reflection image (cropped from a random background image in the same split \ref{fig:synthetic-image-generation}.b). The blended screen with reflection, $B$ is calculated as:
\begin{equation}
    \begin{split}
    Y = 1 - (1 - S) * (1 - R) \\ 
    B = Y * (1 - \alpha) + S * \alpha
    \end{split}
\end{equation}
where $\alpha \in [0, 1]$ determines the intensity of the reflection, as shown in Fig. \ref{fig:synthetic-image-generation}.c.
In the scene background image (Fig. \ref{fig:synthetic-image-generation}.d), a random set of four points was produced by creating a rectangle and applying a random displacement to its corners (to a maximum of half the rectangle height and width). The blended screen with reflection, $B$, is then inserted into a background image, by undergoing a perspective transform defined by the four points as shown in Fig. \ref{fig:synthetic-image-generation}.e. An example of the resulting synthetic image with a screen is shown in Fig. \ref{fig:synthetic-image-generation}.f.
This process is repeated twice, with two different background images, to encourage the model to focus on the echo image rather than on the background. Additionally, the background image, without an inserted echo screen, is also added to the training set to guide the supervision of the classifier branch with predicts the presence of a screen with echo content. A summary of the synthesized data is in Table \ref{tab:synthdata}.

\begin{table}[htb!]
  \caption{Summary of synthetic data}
  \label{tab:synthdata}
  \centering
  \begin{tabular}{llll}
    \hline
    Split       & \# with screen  & \# without screen & total \\
    \hline
    Training    & 23791  & 23791     & 47582 \\
    Validation  & 5764   & 5764      & 11528 \\
    Test        & 2448   & 2448 & 4896\\
    \hline
  \end{tabular}
\end{table}

\subsection{Screen detection model design and training}

We adapted the multi-task UNet architecture proposed in \cite{zhu2022multi} by replacing the saliency prediction branch by a four-channel heatmap prediction followed by a DSNT layer \cite{nibali2018numerical} to localize the four corners of the screen where the standard UNet decoder predicts the four corner heatmaps, and leaving the classification branch to predict the presence of a screen. 
%
%
The multi-task learning process was driven by two losses: a screen corner localization loss, $L_s$ (euclidean distance between predicted and reference points), and a screen visibility classification loss, $L_c$ (classification cross entropy). The loss terms were balanced as follows:
\begin{equation}
L = \frac{L_s}{\sigma^2_s} + \frac{L_c}{\sigma^2_c} + \ln  (\sigma_s +1) + \ln (\sigma_c +1)
\end{equation}
where $\sigma_c$ and $\sigma_s$ are learnable parameters that estimate the uncertainty of $L_c$ and $L_s$ respectively~\cite{liebel2018auxiliary}.

\subsection{Geometric correction and post-processing}

\begin{figure}[htb!]
  \centering
\begin{minipage}[t]{0.26\linewidth}
    \includegraphics[width=\linewidth]{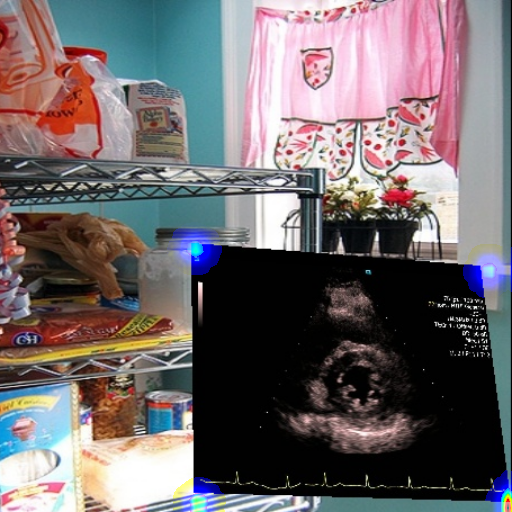}
    \subcaption{Detection}
\end{minipage}
\hfill
\begin{minipage}[t]{0.36\linewidth}
    \includegraphics[width=\linewidth]{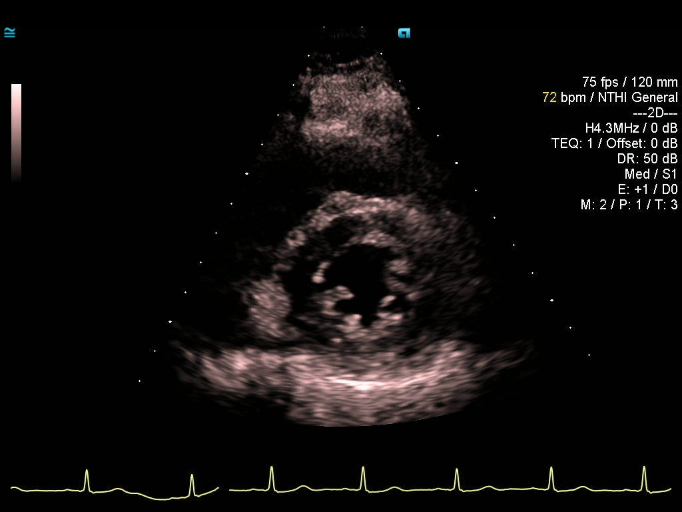}
    \subcaption{Reference}
    \label{fig:homography2}
\end{minipage}
\hfill
\begin{minipage}[t]{0.36\linewidth}
    \includegraphics[width=\linewidth]{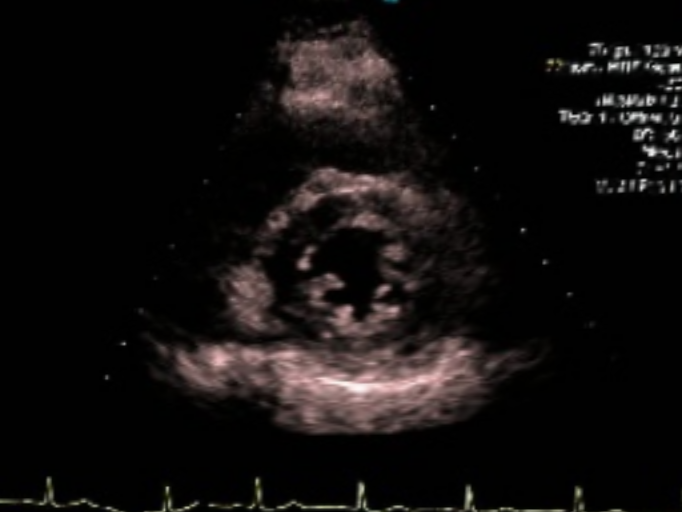}
    \subcaption{Reconstruction}
    \label{fig:homography3}
\end{minipage}
  \caption{Echo frame reconstruction via homgraphy transform of the detected screen. (b) shows the reference frame, compared to the (c), the frame reconstructed form the screen detected in (a).}
  \label{fig:homography}
\end{figure}

Once the coordinates (in pixels) of the four corners of the screen have been detected, the screen content must be compensated for the geometric distortion by the widely known homography transformation \cite{szeliski2022computer}, which is readily available in most computer vision libraries. 
The homography is applied to a user-defined target image grid of $W_t \times H_t$ pixels (here set to $640 \times 480$, commonplace in US industry). An example of the result of the homography transform is shown in Fig \ref{fig:homography}.

After homography has been applied, one may want to post-process the resulting image to reduce perspective artifacts, reflections, and other degradations introduced by the acquisition process. Here we limit ourselves to a very basic normalization process, by which we convert to grayscale, quantize to 256 levels, set the background to black (0 intensity) and clip any values below, and set the maximum value to 255, to finally encode as unsigned int 8 bit (the usual encoding for ultrasound data). To set the background to black we find the background as the most common intensity (after quantization), and linearly stretch intensity values accordingly.

\section{Experiments and results}

We carried out three types of experiments: first, we assessed the performance of the screen localization and detection, using the error norm (in pixels) for the former, and the sensitivity and specificity for the latter, on both the synthetic-test and the real datasets. Second, we measured the similarity between the original ultrasound images and the homography-reconstructed counterparts, using the Structural Similarity Index Measure (SSIM \cite{wang2004image}). Third, we assessed the impact of the capture pipeline when feeding the reconstructed images to a model trained on conventional data, in this case an echo view classifier trained on 2D echo frames.

\subsection{Evaluation of screen detection and localization}

\begin{table*}[h]
  \caption{Screen detection and localization results on synthetic and real data}
  \label{tab:model-synth}
  \centering
  \begin{tabular}{cccccc}
   \hline
   \multicolumn{6}{c}{Synthetic data} \\ 
    N       & 100 & 1000  & 3000 & 10000 & 47582
    \\
    \hline
    Loc. error (px) $\downarrow$ & 2.57 (.27, 30.55) & 0.99 (0.14, 22.10)  &   0.66 (0.09, 18.62)   & 0.43 (0.07 ; 9.01)  & 0.32 (.05, 5.89) \\
    \hline
    Sensitivity $\uparrow$ & .725 (.717, .734) & .877 (.871, .883)  &   .884 (.878, .891)   & .968 (.964, .972) & .991 (.990, .993)\\
    Specificity $\uparrow$  & .816 (.808. .824) & .966 (.963, .971)   &    .981 (.978, .984)   & .993 (.992, .995) & .998 (.997, .999) \\
    Confusion mat.    & $\begin{bmatrix} 1776 & 672 \\ 450 & 1998 \end{bmatrix}$
    & $\begin{bmatrix} 2147 & 301 \\ 82 & 2366 \end{bmatrix}$   & $\begin{bmatrix} 2165 & 283 \\ 47 & 2401 \end{bmatrix}$ & $\begin{bmatrix} 2369 & 79 \\ 16 & 2432 \end{bmatrix}$ & $\begin{bmatrix} 2427 & 21 \\ 5 &  2443 \end{bmatrix}$ \\
    \hline
    \hline
    \multicolumn{6}{c}{Real data} \\ 
    \hline
    Loc. error (px) $\downarrow$ & 4.72 (0.78, 23.03) & 4.76 (1.08 ; 25.57)  &  4.64 (1.36, 21.75)   & 4.21 (1.25, 16.20)  & 4.20 (1.73, 13.90) \\
    \hline
    Sensitivity $\uparrow$ & 1.0 (1.0, 1.0) & .868 (.840, .905)  &    .927 (.910, .962)   & .988 (.987, 1.00) & 0.962 (.950, .975)\\
    Specificity $\uparrow$  & .769 (.728, .812) &  .988 (.987, 1.00)  &  .988 (.987, 1.00)   & .988 (.987, 1.00) & 1.0 (1.0, 1.0) \\
    Confusion mat.    & $\begin{bmatrix} 100 & 0 \\ 23 & 77 \end{bmatrix}$
    & $\begin{bmatrix} 87 & 13 \\ 1 & 99 \end{bmatrix}$   & $\begin{bmatrix} 93 & 7 \\ 1 & 99 \end{bmatrix}$ & $\begin{bmatrix} 99 & 1 \\ 1 & 99 \end{bmatrix}$ & $\begin{bmatrix} 96 & 4 \\ 0 &  100 \end{bmatrix}$ \\
    \hline
  \end{tabular}
\end{table*}

We trained the model 5 times, with an increasing amount of synthetic data (100, 1000, 3000, 10000 and 47582 samples), in all cases for 200 epochs, to assess the impact of adding data into the performance on the test set. The results for synthetic data are shown at the top of Table \ref{tab:model-synth}, which shows the average Euclidean corner localization error (in pixels), and the screen detection error (via the binary sensitivity, specificity and the confusion matrix). The results were bootstrapped 1000 times taking a random 80\% subset of the data each time, providing the 95\% confidence interval and the median (shown in the table as \textit{median (2.5\%, 97.5\%)}). 
The results of the same models on real data are shown at the bottom of Table \ref{tab:model-synth}. In both cases, the pixel localization error decreased monotonically, as expected, when increasing the amount of training data. In the synthetic dataset, the error went (in median) sub-pixel with only 1000 samples in the training set, and the sensitivity, specificity and confusion matrix for screen detection followed a similar trend, with a sensitivity $>0.95$ from 10000 samples and a specificity $> 0.95$ from only 1000 samples in the training set. 
The results on real data followed the same trend but more modestly (4 pixel error, $< 1$\% image size). 

\subsection{Image quality assessment}

\begin{figure}[htb!]
\centering
\offinterlineskip
\begin{overpic}[height=0.27\linewidth]{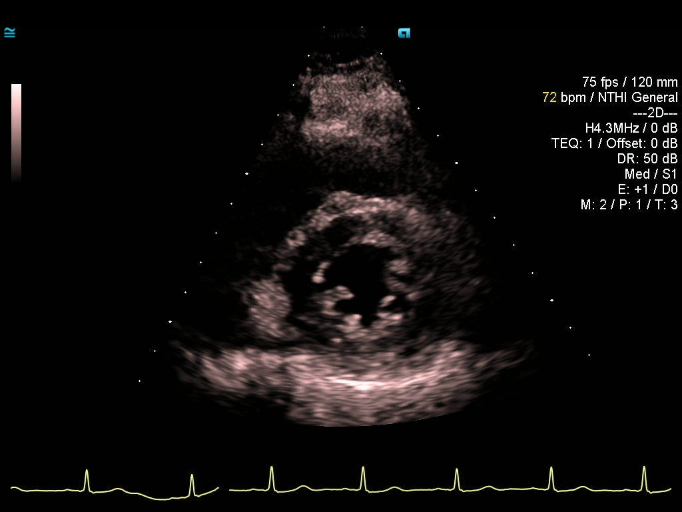}
\put(25,60){\color{white}\textbf{ Original}}
\put(5,5){\color{white}\textbf{\large a}}
\end{overpic}%
\begin{overpic}[height=0.27\linewidth]{figures/00004_capture.png}

\put(22,80){\color{white}\textbf{\contour{black}{Capture}}}
\end{overpic}%
\begin{overpic}[height=0.27\linewidth]{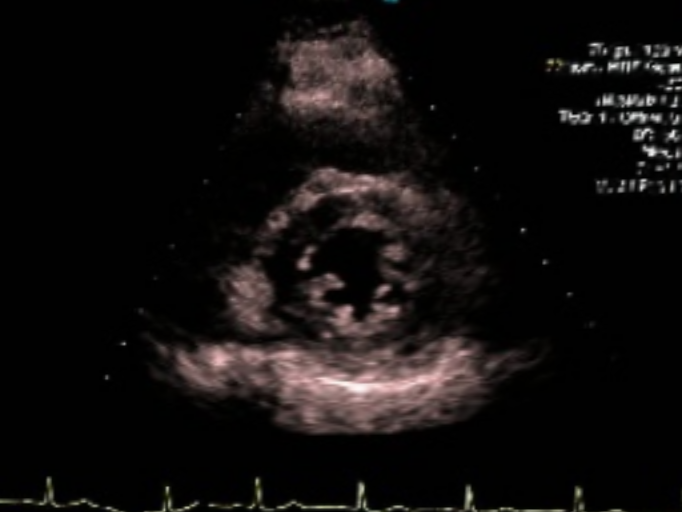}
\put(8,60){\color{white}\textbf{ Reconstructed}}
\put(-2, 5){\color{white}\textbf{ \footnotesize MSE=0.019, SSIM=0.583}}
\end{overpic}\par
\begin{overpic}[height=0.27\linewidth]{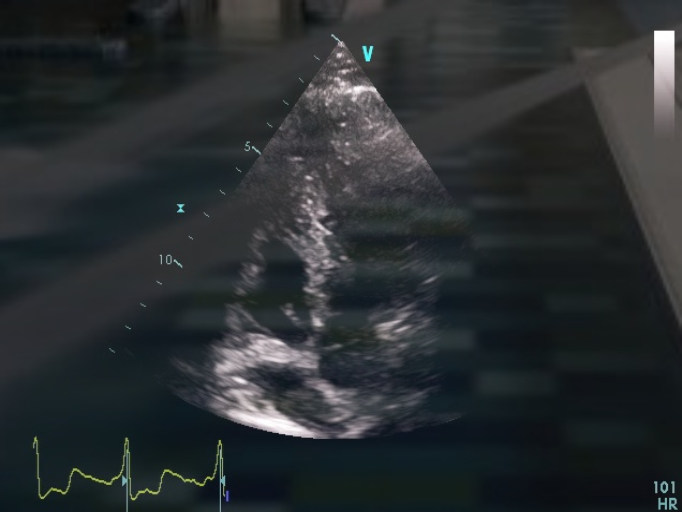}
\put(5,5){\color{white}\textbf{\large b}}
\end{overpic}%
\begin{overpic}[height=0.27\linewidth]{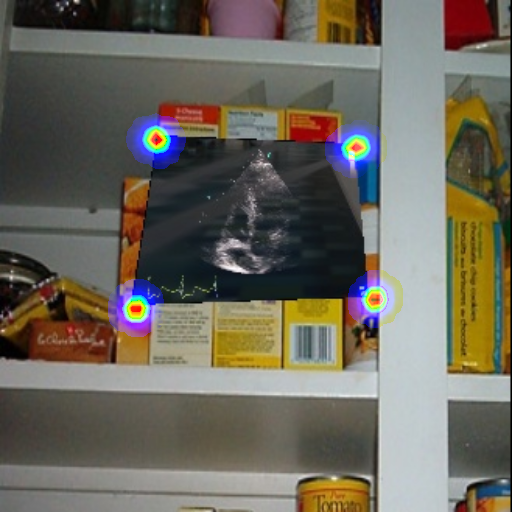}
\end{overpic}%
\begin{overpic}[height=0.27\linewidth]{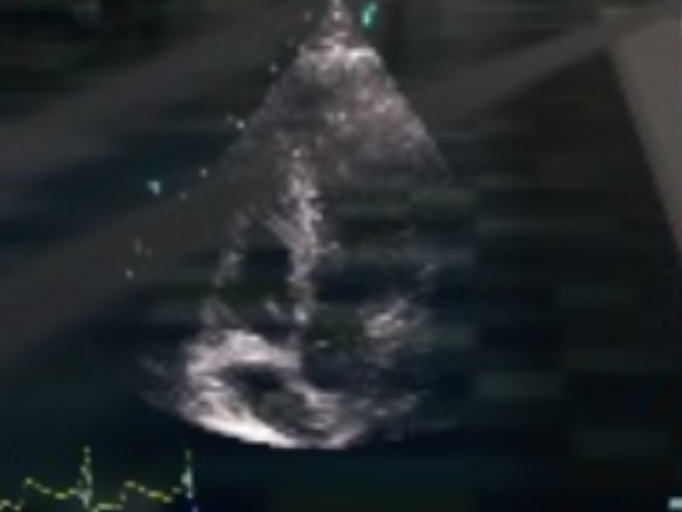}
\put(-2,5){\color{white}\textbf{ \footnotesize MSE=0.014, SSIM=0.650}}
\end{overpic}\par
\begin{overpic}[height=0.27\linewidth]{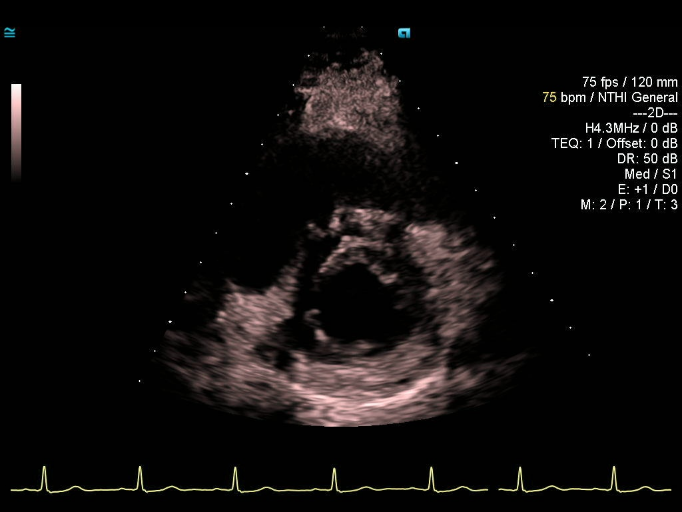}
\put(5,5){\color{white}\textbf{\large c}}
\end{overpic}%
\begin{overpic}[height=0.27\linewidth]{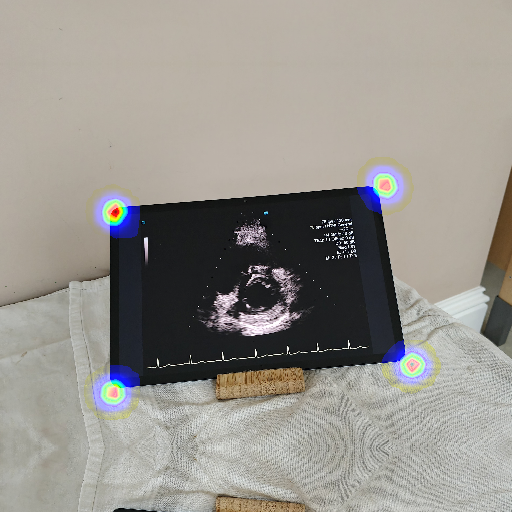}
\end{overpic}%
\begin{overpic}[height=0.27\linewidth]{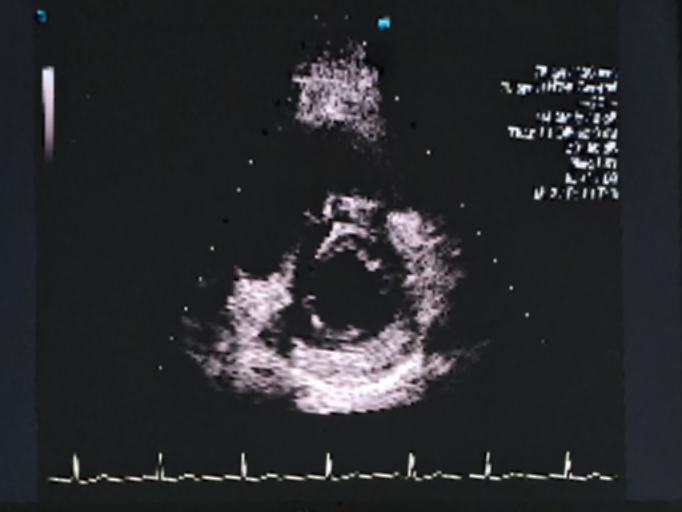}
\put(-2,5){\color{white}\textbf{ \footnotesize MSE=0.038, SSIM=0.037}}
\end{overpic}\par
\begin{overpic}[height=0.27\linewidth]{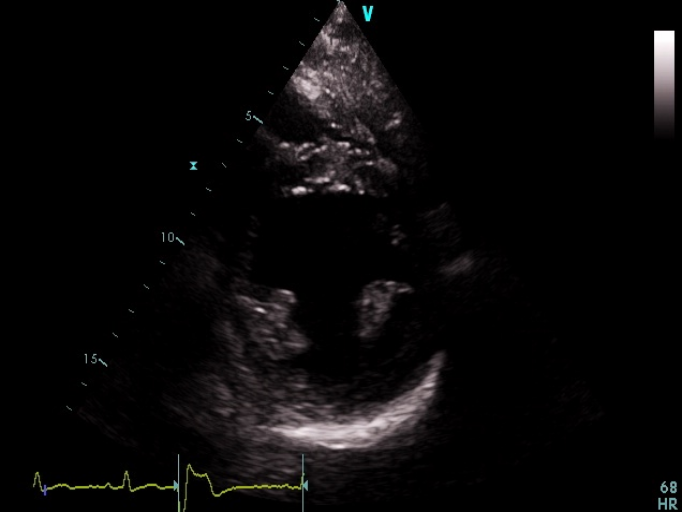}
\put(5,5){\color{white}\textbf{\large d}}
\end{overpic}%
\begin{overpic}[height=0.27\linewidth]{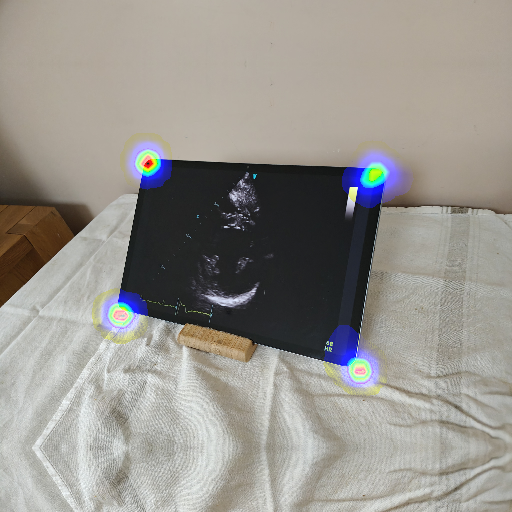}
\end{overpic}%
\begin{overpic}[height=0.27\linewidth]{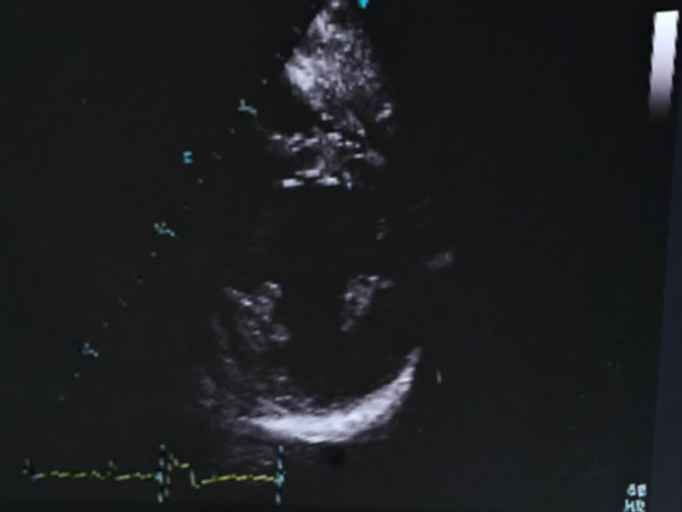}
\put(-2,5){\color{white}\textbf{ \footnotesize MSE=0.013, SSIM=0.085}}
\end{overpic}\par
\caption{Examples of original and reconstructed images, and matching MSE and SSIM values. (a), (b) are from the synthetic dataset, and (c), (d) are from the real dataset.}
\label{fig:image-quality}
\end{figure}

The quality of reconstructed images was measured using the pixel-wise Mean Squared Error (MSE) and the Structural Similarity Index (SSIM \cite{wang2004image}) with respect to the original echo images, providing median values with a 95\% confidence interval. For synthetic images was MSE= $0.01 \ (0.005,  0.040)$ and SSIM= $0.57 \ (0.337, 0.79)$, and for real images MSE= $0.03\ (0.01, 0.08)$ and SSIM= $0.1\ (0.03, 0.28)$. To contextualize these relatively large numbers, a few example pairs of original vs reconstructed images are shown in Fig. \ref{fig:image-quality}.


\subsection{Evaluation in echo view classification}

\begin{figure}[htb!]
  \centering
\begin{minipage}[t]{0.55\linewidth}
    \includegraphics[height=160px,  trim={0, 0, 55, 0}, clip]{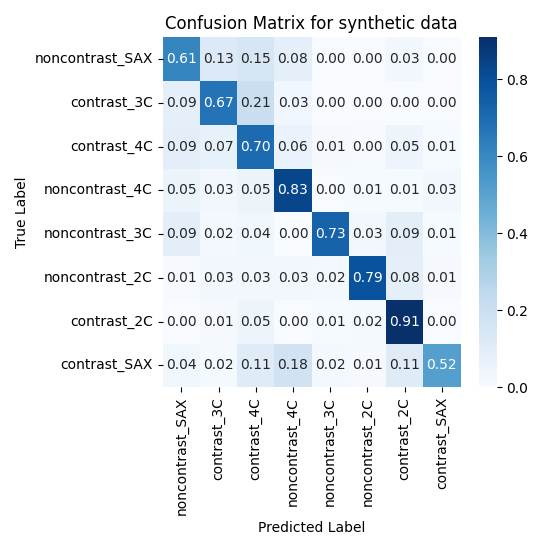}
    \subcaption{Synthetic}
\end{minipage}
\hfill
\begin{minipage}[t]{0.44\linewidth}
    \includegraphics[height=160px, trim={119, 0, 0, 0}, clip]{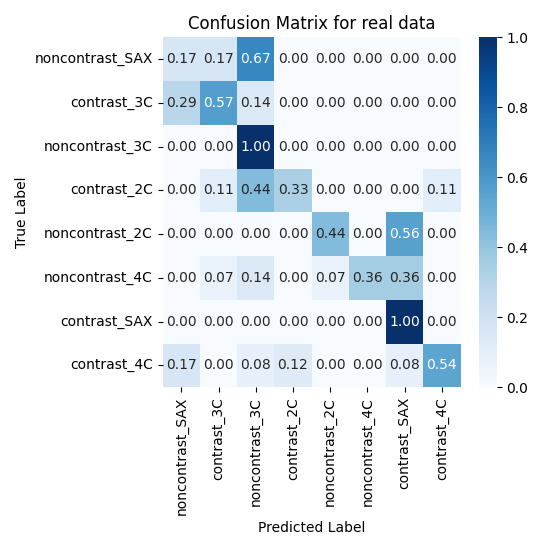}
    \subcaption{Real}
    \label{fig:cm2}
\end{minipage}
  \caption{Confusion matrix for the view classifier (vs using the original images) for the synthetic (left) and the real dataset (right) after removing the 20\% most uncertain samples.}
  \label{fig:cms}
\end{figure}

We run the view classification model from \cite{chartsias2021contrastive} to the reconstructed images and assessed the performance drift with respect to the original images. The balanced accuracy was 0.65 (synthetic) and 0.47 (real). Considering that reflections may  lead to uncertainty and misclassifications, we used the maximum probability in all classes as an uncertainty measure \cite{hendrycks2016}. The balanced accuracy after removing the 20\% and 40\% most uncertain samples increased to 0.72 and  0.79 for synthetic and 0.55 and 0.56 for real data. The corresponding confusion matrices (for 20\% removal) are shown in Fig. \ref{fig:cms}.

\section{Conclusion}

We have presented a methodology generating labeled data for a US screen detection task, and to extract the image content from it, so that it can be plugged into a conventional US image analysis pipeline. Notably, we overcome challenges associated with wide variety of screen shapes and sizes, and potential confusion with other screens (computers, TVs, etc), by detecting ultrasound content rather that screen-shaped objects. The resulting performance, as well as the reconstructed images are promising however we have observed a decline when testing on real images. We hypothesize that this decline can be due to a number of factors including 1) ambiguity in the manual labeling process, 2) uncertainty around the screen frame (especially when it is black as in the real dataset), 3) other sources of image degradation on top of geometric distortion and reflections that we have not modeled. All these will be investigated in future work.



\bibliographystyle{IEEEbib}
\bibliography{bibliography}

\section{Compliance with Ethical Standards}

This research study was conducted retrospectively using human subject data from a proprietary dataset at Ultromics, provided by external collaborators. Ethical approval was obtained by the respective institutional review boards of each collaborator. 

\section{Conflicts of Interest}

All authors are employees of Ultromics Ltd.

\end{document}